\PassOptionsToPackage{table}{xcolor}
\documentclass[10pt,twocolumn,letterpaper]{article}

\usepackage{wacv}

\usepackage{amsfonts}
\usepackage{array}
\usepackage{textcomp}
\usepackage{verbatim}
\usepackage{multirow}
\usepackage{makecell}
\usepackage{colortbl}
\definecolor{wacvblue}{rgb}{0.21,0.49,0.74}
\usepackage[pagebackref,breaklinks,colorlinks,allcolors=wacvblue]{hyperref}
\hypersetup{
  pdftitle={Efficient Real-World Dehazing via Physics-Inspired Global-Local Decoupling},
  pdfauthor={Yifei Qu, Ru Li, Junjie Chen, Jinyuan Wu}
}

\def\wacvPaperID{1673}
\def\confName{WACV}
\def\confYear{2027}

\graphicspath{{imgs/}}

\newcommand{\best}[1]{\textbf{#1}}
\newcommand{\second}[1]{\underline{#1}}

\begin{document}

\title{Efficient Real-World Dehazing via Physics-Inspired Global-Local Decoupling}

\author{Yifei Qu \quad Ru Li$^\dagger$ \quad Junjie Chen \quad Jinyuan Wu\\
Harbin Institute of Technology, China\\
{\tt\small 2023211862@stu.hit.edu.cn, liru@hit.edu.cn, csjunjie.c@gmail.com, 25s130347@stu.hit.edu.cn}}

\maketitle
\begingroup
\renewcommand{\thefootnote}{\fnsymbol{footnote}}
\footnotetext[2]{Corresponding author.}
\endgroup

\begin{abstract}
Real-world single image dehazing is highly ill-posed due to spatially and spectrally varying scattering, while practical deployment demands lightweight and low-latency models. Existing approaches either rely on fragile physical inversion under simplified assumptions or adopt heavy blind architectures unsuitable for edge deployment. To overcome these limitations, we propose \textbf{PGL-Net} (\textbf{P}hysics-Inspired \textbf{G}lobal-\textbf{L}ocal Decoupling Network), a lightweight framework that incorporates physical inductive biases via operator-level emulation, avoiding explicit parameter estimation. It decouples dehazing into global distribution rectification and local structural refinement. A Physics-Inspired Affine Fusion (PAF) module performs globally conditioned alignment across hierarchical skip connections to compensate for haze-induced bias, while a compact Degradation-Aware Modulation (DAM) block adaptively restores spatially and spectrally variant details through dynamic feature modulation. Extensive experiments on multiple real-world benchmarks demonstrate that PGL-Net achieves state-of-the-art restoration quality with significantly reduced complexity. Compared with the recent SOTA SGDN, the Tiny variant (\textbf{PGL-Net-T}) improves PSNR by up to \textbf{2.6\,dB} and consistently enhances downstream object detection accuracy, while achieving over a \textbf{10$\times$ reduction} in inference latency. Code is available at \href{https://github.com/sc-30-bit/PGL-Net}{github.com/sc-30-bit/PGL-Net}.
\end{abstract}

\section{Introduction}
\label{sec:intro}

Single image dehazing aims to recover latent clear scene radiance from observations degraded by atmospheric scattering. 
In real-world applications such as autonomous driving, aerial surveillance, and maritime monitoring, haze severely reduces visibility and suppresses discriminative structures, impairing downstream tasks (e.g., image segmentation~\cite{kirillov2023sam,xie2021segformer}, object detection~\cite{zhao2024detrs,lei2025yolov13}, and tracking~\cite{hu2023track2,hu2024track1}).
Meanwhile, practical deployment on resource-constrained edge platforms demands lightweight and low-latency dehazing networks. Several efficient models have been explored for real-time or compact deployment~\cite{cai2016dehazenet,ren2016mscnn,li2017aod,ren2018gatedfusion,chen2019wacv}, yet achieving strong restoration quality and consistent downstream gains under strict efficiency constraints remains challenging.

\begin{figure}[t]
    \centering
    \includegraphics[width=\linewidth]{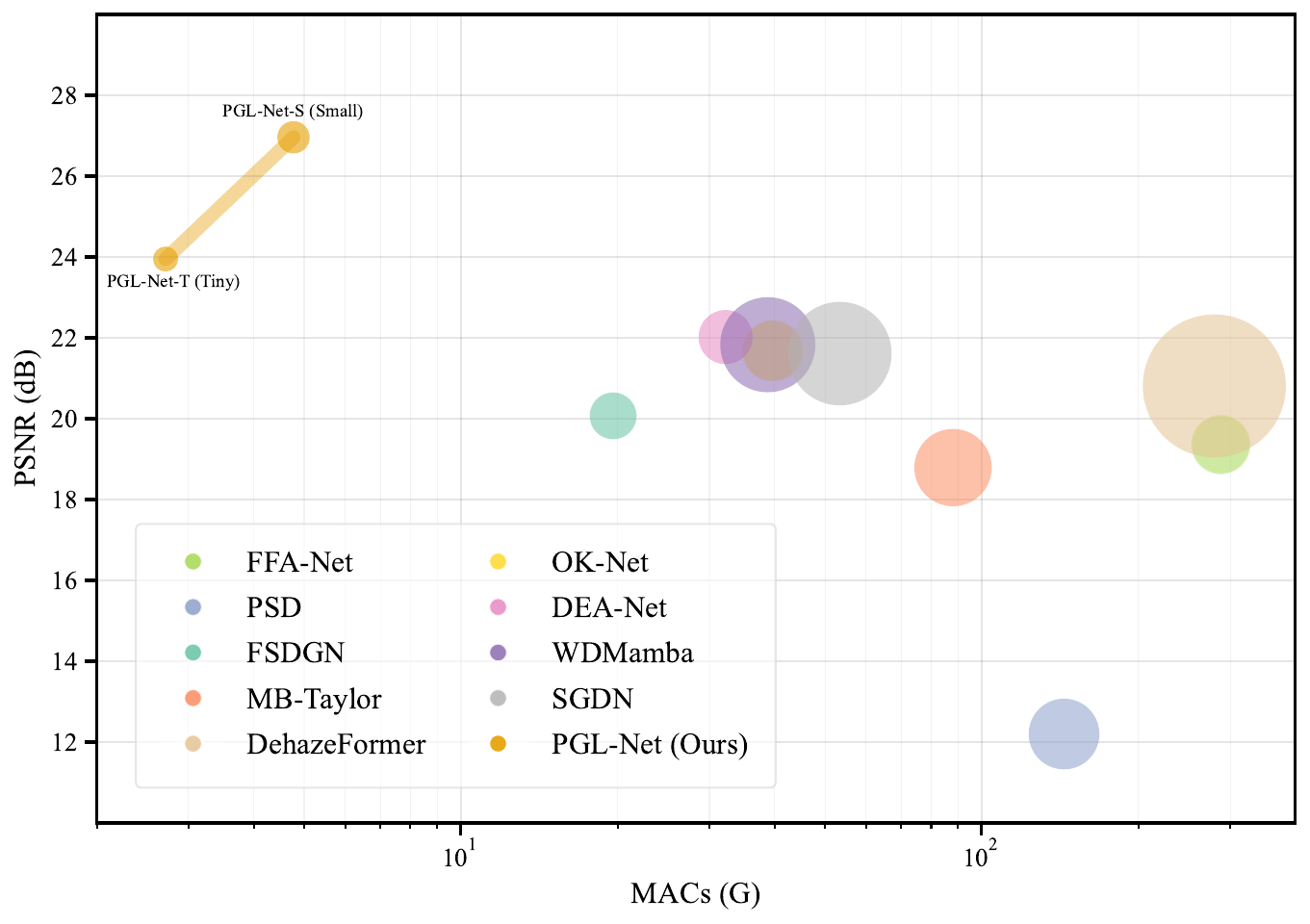}
    \caption{Performance--efficiency comparison on NH-HAZE21. Bubble size denotes the number of parameters, with MACs shown on a logarithmic scale.}
    \label{fig:tradeoff}
\end{figure}

The haze formation process is commonly described by the Atmospheric Scattering Model (ASM)~\cite{mccartney1976ASM1optics,nayar1999ASM2,narasimhan2002ASM3}:
\begin{equation}
    I(x) = J(x)t(x) + A(1-t(x)),
\end{equation}
where $I(x)$ is the observed hazy image, $J(x)$ denotes the latent clear scene radiance, and $A$ represents the global atmospheric light.
The transmission map $t(x)=e^{-\beta d(x)}$ models the portion of scene radiance that reaches the camera, where $\beta$ is the atmospheric scattering coefficient and $d(x)$ denotes scene depth.
Accordingly, haze degradation consists of a multiplicative attenuation term $J(x)t(x)$ and an additive airlight component $A(1-t(x))$~\cite{he2009DCP}.

In real-world scenarios, the ASM becomes 
inaccurate due to wavelength-dependent scattering~\cite{mccartney1976ASM1optics}.
When atmospheric particles are small or objects are distant, shorter wavelengths are scattered more strongly than longer ones, causing the effective scattering coefficient $\beta$ to vary across color channels.
As a result, the transmission can no longer be modeled as a shared scalar but becomes channel-dependent (e.g., $t_r(x), t_g(x), t_b(x)$), introducing additional ambiguities and rendering the standard ASM insufficient for complex real-world scenes.

Existing methods for image dehazing can be roughly divided into prior-driven physical inversion and blind end-to-end restoration. Prior-based and early learning-based approaches usually follow an ``estimate-and-invert'' pipeline, where transmission maps and atmospheric light are explicitly estimated according to simplified atmospheric scattering assumptions. However, such explicit parameter estimation is often fragile in real-world scenarios, since sRGB images have undergone nonlinear camera ISP operations and real haze is frequently accompanied by non-uniform illumination and scene-dependent degradation. As a result, these methods may suffer from residual haze, color distortion, or unstable restoration. In contrast, recent blind end-to-end models avoid explicit physical estimation and learn direct hazy-to-clear mappings with deep architectures, iterative refinement, or generative pipelines. Although effective, their increasing computational complexity limits practical deployment on resource-constrained devices.

To bridge the gap between rigid physical inversion and heavy blind mapping, we propose \textbf{PGL-Net} (\textbf{P}hysics-Inspired \textbf{G}lobal--\textbf{L}ocal Decoupling Network), a lightweight U-shaped architecture that embeds physical inductive bias through operator-level emulation rather than explicit parameter recovery. Instead of directly estimating the transmission $t(x)$ and atmospheric light $A$, PGL-Net preserves the inverse-ASM operator form, namely multiplicative compensation and additive bias correction, and implements it as feature-space affine rectification. Motivated by the observation that haze degradation contains both globally shared atmospheric statistics and spatially varying local degradation, PGL-Net decouples restoration into global feature rectification and local degradation refinement. Specifically, we introduce a Physics-Inspired Affine Fusion (PAF) module to perform globally conditioned channel-wise affine rectification across skip connections, mitigating the haze-induced distribution mismatch between encoder and decoder features. The remaining spatially and spectrally variant degradations are further handled by a lightweight Degradation-Aware Modulation (DAM) block for adaptive local refinement. As shown in Fig.~\ref{fig:tradeoff}, the Tiny variant, PGL-Net-T, achieves a favorable quality--efficiency trade-off on NH-HAZE21~\cite{ancuti2021nh}, while using only 5.8\% of the parameters and 5.0\% of the MACs compared with SGDN~\cite{fang2025sgdn}.

Our main contributions are summarized as follows:
\begin{itemize}
    \item We propose PGL-Net, a lightweight physics-inspired architecture that decouples dehazing into global distribution rectification and local structural refinement for real-world edge deployment.
    \item We introduce PAF, a globally conditioned affine fusion module that implicitly emulates attenuation compensation and airlight suppression to align feature distributions across skip connections.
    \item We design an efficient DAM block for locally adaptive refinement, achieving strong detail restoration with minimal parameter overhead.
    \item Extensive experiments on multiple real-world benchmarks demonstrate state-of-the-art restoration quality and consistent improvements in downstream object detection, validating the practical utility of PGL-Net.
\end{itemize}

\begin{figure*}[!t]
    \centering
    \includegraphics[width=0.8\textwidth]{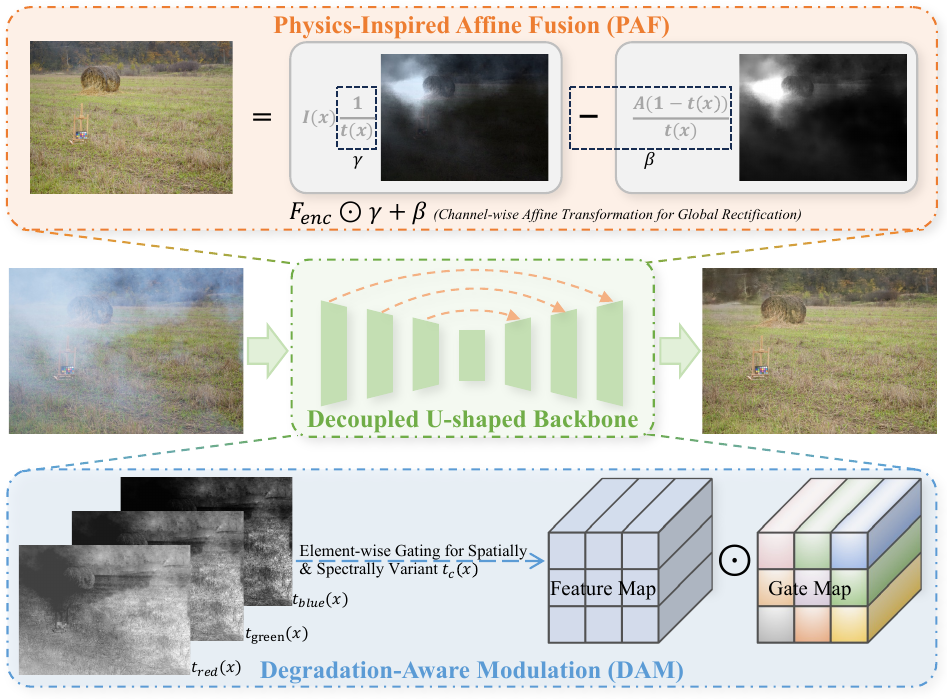}
    \caption{Overview of PGL-Net.
PGL-Net decouples single-image dehazing into global distribution rectification and local structural refinement.
Top: PAF-based skip-connection module. From the inverse atmospheric scattering model, dehazing can be interpreted as a channel-wise affine transformation in feature space, where the \textit{scale} and \textit{shift} correspond to transmission compensation ($\gamma \rightarrow \frac{1}{t(x)}$) and airlight-related bias correction ($\beta$), respectively.
Middle: Decoupled U-shaped backbone. The overall encoder--decoder architecture processes the hazy input and reconstructs the clean output, while skip features are globally rectified before fusion.
Bottom: Degradation-Aware Modulation (DAM). After global rectification, remaining spatially and spectrally variant degradations are handled by stacked DAM blocks, which perform element-wise gating to adaptively restore fine structures under different haze densities.}
\label{fig:overview}
\end{figure*}

\section{Related Work}
\label{sec:related}

\subsection{Single-Image Dehazing}
Early methods~\cite{fattal2008single,tan2008visibility,he2009DCP,zhu2015color3,berman2016nonlocal,bui2017color1} relied on the Atmospheric Scattering Model (ASM) and handcrafted statistical priors to constrain the ill-posed dehazing problem. Representative examples include the Dark Channel Prior~\cite{he2009DCP}, Color Attenuation Prior~\cite{zhu2015color3}, and Non-Local Prior~\cite{berman2016nonlocal}. Despite their effectiveness under specific assumptions, these priors are fragile in complex real-world scenes. Bright objects, wavelength-dependent scattering, and spatially varying haze can lead to color distortion or residual haze.

With the advent of deep learning, early data-driven methods~\cite{cai2016dehazenet,ren2016mscnn,zhang2018dcpdn,li2017aod} estimated ASM parameters such as transmission and atmospheric light. Although more flexible than handcrafted priors, they retained the ``estimate-and-invert'' paradigm and remained sensitive to parameter-estimation errors. Subsequent studies therefore shifted toward blind end-to-end restoration~\cite{liu2019griddehazenet,qin2020ffa,yu2022fsdgn,song2023dehazeformer,chen2024dea,zheng2024mambadehazing,fang2025sgdn}, directly mapping hazy images to clean outputs with increasingly capable architectures.

For real-world haze, further efforts explored domain adaptation and unpaired learning~\cite{chen2021psd,zhao2021refinednet,chen2022cdd,wu2023ridcp,ma2025coa}, while diffusion- and flow-based models introduced alternative generative formulations~\cite{yang2024diffusion1,wang2025diffusion2,shin2025hazeflow,xin2026fsflow}. These approaches can incur substantial computation or iterative inference. Meanwhile, existing dehazing architectures primarily emphasize backbone feature extraction, with less attention paid to adaptive fusion between encoder and decoder representations produced at different restoration stages.

\subsection{Feature Modulation and Adaptive Fusion}
Feature-wise modulation has been widely explored in conditional representation learning.
FiLM~\cite{perez2018film} applies channel-wise scaling and shifting conditioned on an auxiliary representation,
while AdaIN~\cite{huang2017adain} performs feature normalization followed by channel-wise statistics matching.
Attention mechanisms such as SE~\cite{hu2018senet} and
CBAM~\cite{woo2018cbam} are widely used for feature recalibration,
while adaptive encoder--decoder fusion has been explored through
attention gating and selective-kernel mechanisms
~\cite{oktay2018AG,song2022gUNet,song2023dehazeformer}.

The proposed PAF shares the generic affine form of FiLM-like modulation, but the affine operator itself is not our claimed novelty. Instead, PAF is designed specifically for encoder--decoder skip rectification in dehazing, with its modulation jointly conditioned on both branches. Unlike AdaIN, PAF performs neither feature normalization nor explicit statistics matching. These distinctions are further discussed in Sec.~\ref{sec:paf}.

\section{Method}

\label{sec:method}

\begin{figure*}[!t]
    \centering
    \includegraphics[width=\textwidth]{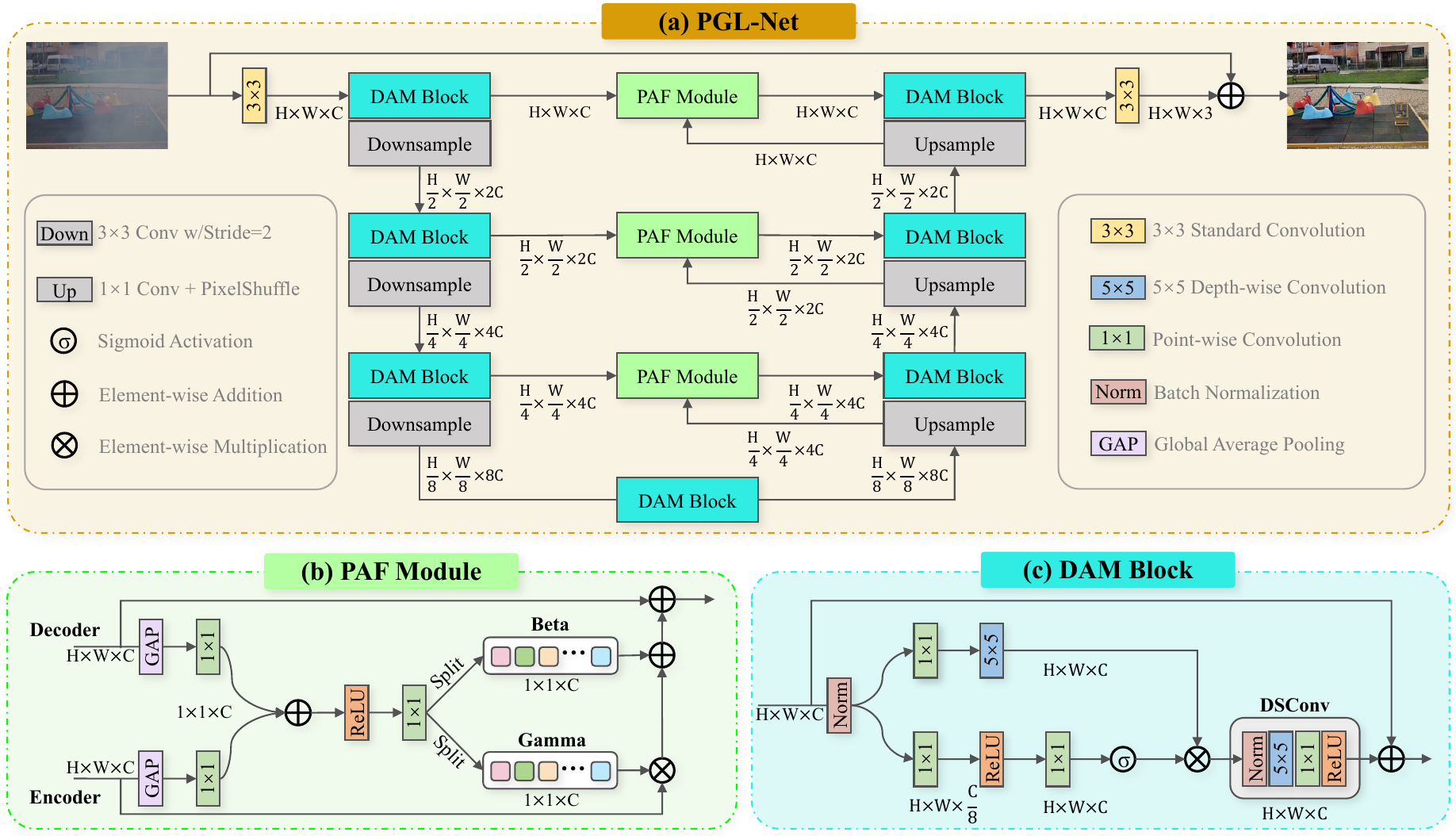}
    \caption{Detailed architecture of PGL-Net.}
    \label{fig:architecture}
\end{figure*}

PGL-Net is a lightweight physics-inspired dehazing framework that decouples restoration into 
global distribution rectification and local structural refinement. 
Fig.~\ref{fig:overview} illustrates the overall global--local pipeline, 
while Fig.~\ref{fig:architecture} shows the detailed architecture.
The Physics-Inspired Affine Fusion (PAF) module performs globally conditioned feature rectification at skip connections, 
interpreting dehazing as a channel-wise affine transformation in feature space. 
Complementing this global correction, local refinement is conducted within the backbone using stacked 
Degradation-Aware Modulation (DAM) blocks, 
which adaptively recalibrate spatially and spectrally variant features. 
Together, these two components form a unified operator-level framework that embeds physical inductive biases with efficient feature modulation.

\subsection{Physics-Inspired Affine Fusion}
\label{sec:paf}

Skip connections in U-shaped dehazing networks typically fuse features by direct concatenation or summation~\cite{ren2018gatedfusion,yang2019wavelet,zhou2023breaking,qiu2023mb,zheng2024mambadehazing,jin2025mbv2}, implicitly assuming compatible feature distributions between the encoder and decoder.
However, encoder and decoder representations are produced at different stages of restoration and may exhibit a substantial distribution discrepancy.
Such a discrepancy can limit the effectiveness of direct fusion. To bridge this gap, we propose the Physics-Inspired Affine Fusion (PAF) module, which explicitly aligns these features using a physically grounded transformation before fusion.

Derived from the inverse atmospheric scattering model, dehazing can be approximated as an affine rectification:
\begin{equation}
    J(x) \approx I(x)\cdot \underbrace{\frac{1}{t(x)}}_{\gamma}
    \;+\;
    \underbrace{A\left(1 - \frac{1}{t(x)}\right)}_{\beta},
\end{equation}
where the scale factor $\gamma$ compensates for attenuation and the shift factor $\beta$ accounts for airlight-induced bias.
Although the transmission map $t(x)$ varies spatially and spectrally, the atmospheric light $A$ is typically modeled as a scene-level constant, indicating that the dominant rectification component is globally conditioned.
Accordingly, PAF performs channel-wise affine alignment with parameters estimated via Global Average Pooling (GAP), embedding this global prior into the network architecture.

As illustrated in Fig.~\ref{fig:architecture}(b), given decoder and encoder features $F_{\mathrm{dec}}$ and $F_{\mathrm{enc}}$, their global descriptors are first extracted via GAP:
\begin{equation}
    s_{\mathrm{dec}}=\mathrm{GAP}(F_{\mathrm{dec}}),\quad
    s_{\mathrm{enc}}=\mathrm{GAP}(F_{\mathrm{enc}}),
\end{equation}
where $s_{\mathrm{dec}}, s_{\mathrm{enc}}\in\mathbb{R}^{C\times1\times1}$.
To ensure that the affine parameters are mutually aware of both the hazy input and the progressive clean reconstruction, these descriptors are projected and aggregated to form a joint representation $z$:
\begin{equation}
    z = \text{ReLU}\!\left(\text{PWConv}(s_{\mathrm{dec}})
    + \text{PWConv}(s_{\mathrm{enc}})\right),
\end{equation}
where $\text{PWConv}$ denotes a point-wise $1\times1$ convolution used for cross-domain feature alignment.
Subsequently, the channel-wise affine parameters are predicted from $z$:
\begin{equation}
    \gamma, \beta = \text{Split}(\text{PWConv}(z)), 
\end{equation}
where $\gamma,\beta\in\mathbb{R}^{C\times1\times1}$. Here, $\text{PWConv}$ efficiently projects the aggregated descriptor into $2C$ channels, which are then evenly divided along the channel dimension by the $\text{Split}$ operation to yield the scale $\gamma$ and shift $\beta$.

Finally, the rectified skip feature is fused with the decoder feature:
\begin{equation}
    F_{\mathrm{out}} = F_{\mathrm{dec}} + (F_{\mathrm{enc}}\odot\gamma + \beta).
\end{equation}

Although $\gamma$ and $\beta$ are globally generated channel-wise factors, the correction induced by PAF remains spatially varying because it is applied to the encoder feature map.
For channel $c$, this correction is:
\begin{equation}
    \Delta F_c(x)=F_{\mathrm{enc},c}(x)(\gamma_c-1)+\beta_c .
\end{equation}

Thus, regions with different haze densities, textures, and local structures can receive different correction responses through $F_{\mathrm{enc},c}(x)$. In this way, PAF provides globally conditioned yet content-dependent feature rectification, effectively alleviating the encoder--decoder distribution mismatch with negligible computational overhead. This global rectification creates a stable foundation for the subsequent Degradation-Aware Modulation blocks, which further handle localized and fine-grained degradations.

PAF shares the generic channel-wise affine form with FiLM-like modulation~\cite{perez2018film}, but differs in its conditioning strategy and functional role.
FiLM predicts affine parameters from a conditioning representation to modulate a feature stream, whereas PAF jointly derives $\gamma$ and $\beta$ from both encoder and decoder descriptors and applies them 
to rectify the encoder skip feature before fusion.
Unlike AdaIN~\cite{huang2017adain}, PAF performs neither feature normalization nor explicit channel-wise statistics matching.
Instead, PAF uses an ASM-motivated scale--shift structure as a jointly conditioned skip-rectification mechanism for dehazing.

\subsection{Degradation-Aware Modulation}
\label{sec:dam}

To address the remaining spatially and spectrally variant degradations after global rectification by PAF, we introduce the Degradation-Aware Modulation (DAM) block. 
Operating in a fully convolutional manner, DAM performs residual feature recalibration with stable normalization and lightweight modulation to 
handle local haze variations.

As illustrated in Fig.~\ref{fig:architecture}(c), given an input tensor $F_{in} \in \mathbb{R}^{H\times W \times C}$, we first normalize the features:
\begin{equation}
    \hat{F} = \text{BN}(F_{in}),
\end{equation}
where Batch Normalization (BN)~\cite{ioffe2015BN} is employed in this fully convolutional setting. 
Compared to Layer Normalization (LN)~\cite{ba2016LN}, which is commonly adopted in Transformer-based restoration models~\cite{guo2022PAFsimilar,qiu2023mb,song2023dehazeformer,zhang2024pcs,jin2025mbv2}, BN 
preserves local spatial statistics that are critical for element-wise modulation under spatially varying haze. 
Furthermore, its fixed population statistics can be fused into adjacent convolutions during inference, introducing negligible runtime overhead.

The normalized representation is then decomposed into two complementary branches. 
The value branch extracts local structural cues:
\begin{equation}
    V = \text{DWConv}_{5\times5}(\text{PWConv}(\hat{F})),
\end{equation}
where the $5\times5$ depth-wise convolution enlarges the receptive field while maintaining low complexity.

In parallel, a compact bottleneck pathway predicts a degradation-aware modulation map:
\begin{equation}
    M \! = \! \text{Sigmoid}\!\left( 
        \text{PWConv}^{\uparrow} 
        \big( \text{ReLU}(\text{PWConv}^{\downarrow}(\hat{F})) \big) 
    \right),
\end{equation}
with a reduction ratio $r=8$. 
This bottleneck design ($C \rightarrow C/8 \rightarrow C$) enables expressive channel-wise responses under varying haze densities while strictly controlling parameters.

Finally, local features are adaptively modulated and aggregated in a residual manner:
\begin{equation}
    F_{out} = \text{DSConv}(V \odot M) + F_{in}.
\end{equation}

The element-wise interaction selectively recalibrates spatial responses, and the subsequent depthwise-separable convolution (BN $\rightarrow$ DWConv$_{5\times5}$ $\rightarrow$ PWConv $\rightarrow$ ReLU) consolidates the refined features.

Compared with heavy gating mechanisms that rely on channel expansion or full-rank projections~\cite{ren2018gatedfusion,tu2022maxim,chen2022NAFNet,zamir2022restormer,song2022gUNet}, DAM achieves a 
balance between modulation flexibility and efficiency under strict computational constraints. 
The gating parameters are reduced to approximately 25\% of gUNet~\cite{song2022gUNet} and just 4\% of Restormer~\cite{zamir2022restormer}, while preserving strong local restoration capability suitable for real-time edge deployment.

\subsection{Objective Function}
\label{sec:loss}

To jointly enforce spatial fidelity and spectral consistency, we train PGL-Net with a composite loss:
\begin{equation}
\mathcal{L}
=
\|\hat{J}-J\|_1
+
\lambda
\left(
\|\Re(\Delta_{\mathcal{F}})\|_1
+
\|\Im(\Delta_{\mathcal{F}})\|_1
\right),
\label{eq:loss}
\end{equation}
where $\hat{J}$ and $J$ denote the restored and ground-truth images, respectively, and 
$\Delta_{\mathcal{F}}=\mathcal{F}(\hat{J})-\mathcal{F}(J)$ denotes their frequency-domain discrepancy after Fast Fourier Transform (FFT).
The first term enforces pixel-wise reconstruction, while the second term constrains spectral consistency by penalizing the real and imaginary components of $\Delta_{\mathcal{F}}$.
This direct complex-domain constraint provides stable optimization while preserving both structural and frequency information.
The weight $\lambda$ is set to $0.1$.


\section{Experiments}

In this section, we conduct comprehensive experiments to verify the effectiveness of PGL-Net, assessing both restoration quality and downstream utility (e.g., object detection). 

\begin{table*}[t!]
\centering
\caption{Quantitative comparison on the RRSHID dataset. $\downarrow$ indicates that lower values are better, while $\uparrow$ indicates that higher values are better. The best results are highlighted in \textbf{bold}, and the second-best results are \underline{underlined}. Methods marked with `*' are cited from public results. Other results are obtained by retraining the models using their official implementations with default configurations. MACs are measured on 256$\times$256 images.}
\label{tab:rrshid_comparison}
\resizebox{\textwidth}{!}{
\begin{tabular}{l c ccc ccc ccc ccc cc}
\toprule
\multirow{2}{*}{\textbf{Methods}} & 
\multirow{2}{*}{\textbf{Venue}} & 
\multicolumn{3}{c}{\textbf{RRSHID-thin}} & 
\multicolumn{3}{c}{\textbf{RRSHID-moderate}} & 
\multicolumn{3}{c}{\textbf{RRSHID-thick}} & 
\multicolumn{3}{c}{\textbf{RRSHID-average}} & 
\multicolumn{2}{c}{\textbf{Overhead}} \\
\cmidrule(lr){3-5} \cmidrule(lr){6-8} \cmidrule(lr){9-11} \cmidrule(lr){12-14} \cmidrule(lr){15-16}
 & & PSNR$\uparrow$ & SSIM$\uparrow$ & LPIPS$\downarrow$ 
   & PSNR$\uparrow$ & SSIM$\uparrow$ & LPIPS$\downarrow$ 
   & PSNR$\uparrow$ & SSIM$\uparrow$ & LPIPS$\downarrow$ 
   & PSNR$\uparrow$ & SSIM$\uparrow$ & LPIPS$\downarrow$ 
   & Params & MACs \\
\midrule
DCP\cite{he2009DCP} & CVPR'09 
& 18.46 & 0.4564 & 0.4851 
& 17.80 & 0.4856 & 0.4700 
& 18.39 & 0.4843 & 0.4996 
& 18.22 & 0.4754 & 0.4849 
& -- & -- \\

GridDehazeNet*\cite{liu2019griddehazenet} & ICCV'19 
& 22.77 & 0.6145 & 0.4123 
& 22.62 & 0.6468 & \second{0.3833} 
& 23.96 & 0.7112 & 0.3947 
& 23.12 & 0.6575 & 0.3968 
& \second{0.96M} & 21.49G \\

FFA-Net*\cite{qin2020ffa} & AAAI'20 
& 17.08 & 0.4452 & 0.5761 
& 17.40 & 0.5385 & 0.5450 
& 16.71 & 0.4792 & 0.5573 
& 17.06 & 0.4876 & 0.5595 
& 4.45M & 287.8G \\

FSDGN\cite{yu2022fsdgn} & ECCV'22 
& 22.73 & 0.6059 & 0.4164 
& 23.22 & 0.6603 & 0.4092 
& 24.24 & 0.7177 & 0.4088 
& 23.35 & 0.6611 & 0.4109 
& 2.73M & 19.60G \\

SCANet*\cite{guo2023scanet} & CVPRW'23 
& 18.37 & 0.4718 & 0.4827 
& 18.11 & 0.0538 & 0.4962 
& 19.07 & 0.5966 & 0.4734 
& 18.52 & 0.3741 & 0.4841 
& 2.40M & 17.66G \\

Trinity-Net*\cite{chi2023trinity} & TGRS'23 
& 20.51 & 0.5728 & 0.4578 
& 22.46 & 0.5728 & 0.4314 
& 24.11 & 0.7234 & 0.4206 
& 22.36 & 0.6230 & 0.4366 
& 22.16M & 21.63G \\

DehazeFormer\cite{song2023dehazeformer} & TIP'23 
& 22.99 & 0.6202 & 0.4265 
& 23.63 & 0.6714 & 0.4171 
& 24.61 & 0.7263 & 0.4119 
& 23.71 & 0.6724 & 0.4181 
& 25.44M & 279.7G \\

OKNet\cite{cui2024oknet} & AAAI'24 
& 22.96 & 0.6140 & 0.4349 
& 23.47 & 0.6682 & 0.4160 
& 24.61 & 0.7258 & 0.4063 
& 23.63 & 0.6690 & 0.4183 
& 4.72M & 39.67G \\

4KDehazing*\cite{xiao20244k} & SigPro'24 
& 22.83 & 0.6177 & 0.4352 
& 22.47 & 0.6505 & 0.4590 
& 22.55 & 0.6912 & 0.4754 
& 22.62 & 0.6531 & 0.4565 
& 34.55M & 103.9G \\

PCSformer*\cite{zhang2024pcs} & TGRS'24 
& 21.83 & 0.5427 & 0.4963 
& 22.09 & 0.5984 & 0.5021 
& 23.71 & 0.6547 & 0.4996 
& 22.54 & 0.5996 & 0.4993 
& 3.73M & 27.66G \\

PhDNet*\cite{lihe2024phdnet} & Inffus'24 
& 22.64 & 0.6054 & 0.4658 
& 22.92 & 0.6448 & 0.4019 
& 24.28 & 0.6996 & 0.3993 
& 23.28 & 0.6499 & 0.4223 
& 10.03M & 33.14G \\

SGDN\cite{fang2025sgdn} & AAAI'25 
& 23.32 & 0.6374 & 0.3987 
& 23.86 & 0.6793 & 0.3950 
& 25.09 & 0.7345 & \best{0.3923} 
& 24.03 & 0.6826 & 0.3952 
& 13.32M & 53.40G \\

MCAF-Net*\cite{zhu2025RRSHID} & TGRS'25 
& 23.32 & 0.6236 & 0.4023 
& 23.60 & 0.6583 & \best{0.3799} 
& 25.40 & 0.7221 & 0.3942 
& 24.11 & 0.6680 & \second{0.3921} 
& -- & --\\

\rowcolor{gray!15} 
\textbf{PGL-Net-T} & - 
& \second{24.22} & \second{0.6676} & \second{0.3895} 
& \second{24.77} & \second{0.7024} & 0.4030 
& \second{25.65} & \second{0.7573} & 0.3973 
& \second{24.85} & \second{0.7074} & 0.3982 
& \best{0.78M} & \best{2.71G} \\

\rowcolor{gray!15} 
\textbf{PGL-Net-S} & - 
& \best{24.58} & \best{0.6817} & \best{0.3788} 
& \best{25.06} & \best{0.7082} & 0.3977 
& \best{25.69} & \best{0.7614} & \second{0.3925} 
& \best{25.10} & \best{0.7149} & \best{0.3917} 
& 1.30M & \second{4.77G} \\
\bottomrule
\end{tabular}
}
\end{table*}

\subsection{Experimental Setup}
\label{sec:datasets}
\noindent\textbf{Datasets.}
We evaluate PGL-Net on three paired real-world benchmarks: RRSHID~\cite{zhu2025RRSHID}
for remote-sensing imagery, RW$^2$AH~\cite{fang2025sgdn} for well-aligned webcam
observations, and NH-HAZE21~\cite{ancuti2021nh} for spatially non-homogeneous
haze. For unpaired evaluation,
we assess RTTS and URHI~\cite{li2018rttsAndReside} using no-reference metrics.
RTTS is also used to evaluate object detection after dehazing. Detailed dataset
statistics and splits are provided in the \mbox{supplementary material}.

\noindent\textbf{Implementation Details.}
PGL-Net-T and PGL-Net-S use block repetition factors $M=2$ and $M=4$, respectively, with a base channel width of $N=24$ and a depthwise kernel size of $5$. Models are optimized using AdamW on random $256\times256$ crops and trained on a single NVIDIA RTX 3090 GPU. Additional implementation details are provided in the supplementary material.

\noindent\textbf{Evaluation Metrics.}
For paired datasets, we report PSNR, SSIM~\cite{wang2004ssim}, and LPIPS~\cite{zhang2018LPIPS}. For unpaired RTTS and URHI, we report NIQE~\cite{mittal2012NIQE}, BRISQUE~\cite{mittal2012brisque}, and FADE~\cite{choi2015fade}. Detection on RTTS is evaluated using per-class AP and mAP$_{50:95}$. Latency is measured on $512\times512$ inputs using
PyTorch FP32 on a single NVIDIA RTX 3090 GPU.

\begin{figure*}[t!]
  \centering
  \includegraphics[width=0.95\linewidth, trim=0cm 0cm 0cm 0cm, clip]{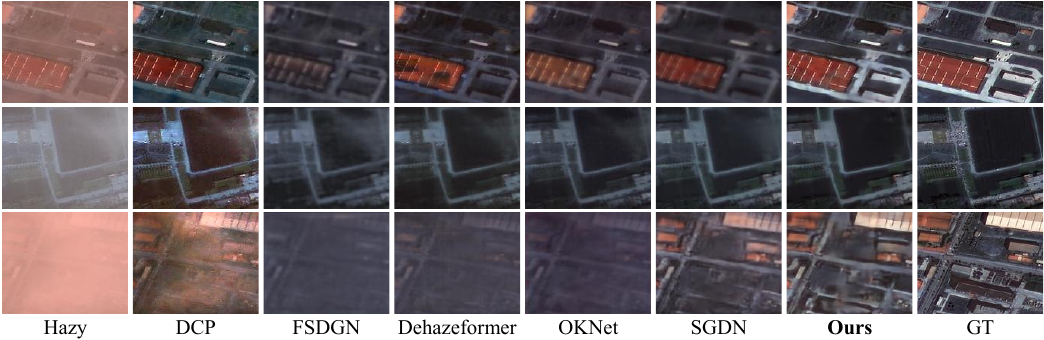}
  \caption{Visual comparison on the RRSHID dataset. Zoom in for a better view.}
  \label{fig:qualitative_rrshid}
\end{figure*}

\begin{figure*}[t!]
  \centering
  \includegraphics[width=0.95\linewidth, trim=0cm 0cm 0cm 0cm, clip]{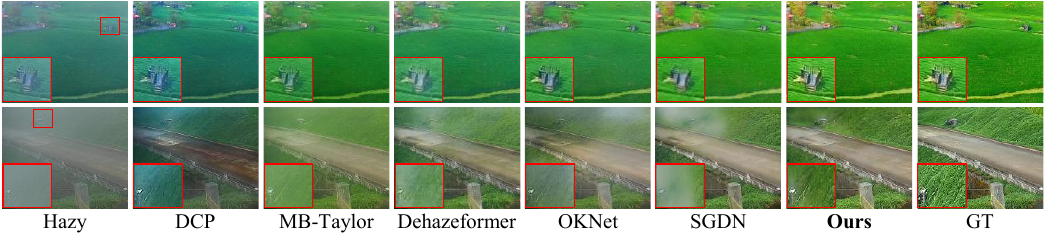}
  \caption{Visual comparison on the RW$^2$AH dataset. Zoom in for a better view.}
  \label{fig:qualitative_rw2ah}
\end{figure*}

\begin{table}[t]
\centering
\caption{Quantitative Comparison on the RW$^2$AH Dataset}
\label{tab:rw2ah_comparison}
\setlength{\tabcolsep}{2.4pt}
\renewcommand{\arraystretch}{1.08}
\resizebox{\columnwidth}{!}{
\begin{tabular}{l c c c c c c}
\toprule
\textbf{Methods} & \textbf{Venue} &
\textbf{PSNR$\uparrow$} & \textbf{SSIM$\uparrow$} &
\textbf{LPIPS$\downarrow$} &
\textbf{Params} & \textbf{MACs} \\
\midrule
DCP~\cite{he2009DCP} & CVPR'09 
& 16.04 & 0.4961 & 0.3363 & -- & -- \\
PSD*~\cite{chen2021psd} & CVPR'21 
& 16.95 & 0.4480 & -- & 6.21M & 143.91G \\
RefineDNet*~\cite{zhao2021refinednet} & TIP'21 
& 16.62 & 0.5170 & -- & 65.78M & 276.81G \\
YOLY*~\cite{li2021yoly} & IJCV'21 
& 15.40 & 0.4510 & -- & \second{1.24M} & 303.90G \\
CDD-GAN*~\cite{chen2022cdd} & ECCV'22 
& 17.78 & 0.6010 & -- & 29.27M & 279.44G \\
FSDGN*~\cite{yu2022fsdgn} & ECCV'22 
& 18.57 & 0.4300 & -- & 2.73M & 19.60G \\
C$^{2}$PNet*~\cite{zheng2023c2p} & CVPR'23 
& 17.48 & 0.4720 & -- & 7.17M & 460.95G \\
MB-Taylor~\cite{qiu2023mb} & ICCV'23 
& 20.33 & 0.5732 & 0.3386 & 7.43M & 88.10G \\
DehazeFormer~\cite{song2023dehazeformer} & TIP'23 
& 21.50 & 0.5967 & 0.3316 & 25.44M & 279.7G \\
OKNet~\cite{cui2024oknet} & AAAI'24 
& 20.76 & 0.5859 & 0.3411 & 4.72M & 39.67G \\
DEA-Net*~\cite{chen2024dea} & TIP'24 
& 21.14 & 0.5740 & -- & 3.65M & 32.23G \\
DCMPNet*~\cite{zhang2024dcmpnet} & CVPR'24 
& 20.13 & 0.5870 & -- & 7.16M & 62.89G \\
NSDNet*~\cite{fan2025nonaligned} & TCSVT'25 
& 21.39 & 0.6190 & -- & 11.38M & 56.86G \\
SGDN~\cite{fang2025sgdn} & AAAI'25 
& 21.89 & 0.6090 & 0.3258 & 13.32M & 53.40G \\
\midrule
\rowcolor{gray!15}
\textbf{PGL-Net-T} & -- 
& \second{22.60} & \second{0.6468} & \second{0.3045} & \best{0.78M} & \best{2.71G} \\
\rowcolor{gray!15}
\textbf{PGL-Net-S} & -- 
& \best{22.82} & \best{0.6544} & \best{0.3011} & 1.30M & \second{4.77G} \\
\bottomrule
\end{tabular}
}
\end{table}

\begin{table}[t]
\centering
\caption{Quantitative comparison on the NH-HAZE21 dataset.}
\label{tab:nhhaze21_comparison}
\setlength{\tabcolsep}{2.4pt}
\renewcommand{\arraystretch}{1.08}
\resizebox{\columnwidth}{!}{
\begin{tabular}{l c ccc cc}
\toprule
\textbf{Methods} & \textbf{Venue} & \textbf{PSNR$\uparrow$} &
\textbf{SSIM$\uparrow$} & \textbf{LPIPS$\downarrow$} &
\textbf{Params} & \textbf{MACs} \\
\midrule
DCP~\cite{he2009DCP} & CVPR'09 & 11.77 & 0.6207 & 0.3371 & -- & -- \\
FFA-Net~\cite{qin2020ffa} & AAAI'20 & 19.36 & 0.7729 & 0.2599 & 4.45M & 287.8G \\
PSD~\cite{chen2021psd} & CVPR'21 & 12.20 & 0.6468 & 0.3668 & 6.21M & 143.91G \\
YOLY~\cite{li2021yoly} & IJCV'21 & 12.22 & 0.3774 & 0.5491 & 1.24M & 303.90G \\
FSDGN~\cite{yu2022fsdgn} & ECCV'22 & 20.07 & 0.7850 & 0.2519 & 2.73M & 19.60G \\
MB-Taylor~\cite{qiu2023mb} & ICCV'23 & 18.79 & 0.7425 & 0.3156 & 7.43M & 88.10G \\
DehazeFormer~\cite{song2023dehazeformer} & TIP'23 & 20.81 & 0.7809 & 0.2616 & 25.44M & 279.7G \\
OKNet~\cite{cui2024oknet} & AAAI'24 & 21.68 & 0.8183 & 0.2334 & 4.72M & 39.67G \\
DEA-Net~\cite{chen2024dea} & TIP'24 & 22.02 & 0.7963 & 0.2057 & 3.65M & 32.23G \\
WDMamba~\cite{sun2025wdmamba} & TCSVT'25 & 21.83 & 0.8096 & 0.2145 & 11.25M & 38.84G \\
SGDN~\cite{fang2025sgdn} & AAAI'25 & 21.61 & 0.8014 & 0.2427 & 13.32M & 53.40G \\
\midrule
\rowcolor{gray!15}
\textbf{PGL-Net-T} & -- & \second{23.95} & \second{0.8378} & \second{0.2027} & \best{0.78M} & \best{2.71G} \\
\rowcolor{gray!15}
\textbf{PGL-Net-S} & -- & \best{26.96} & \best{0.8577} & \best{0.1886} & \second{1.30M} & \second{4.77G} \\
\bottomrule
\end{tabular}
}
\end{table}

\begin{table}[t]
\centering
\caption{No-Reference Evaluation on RTTS and URHI}
\label{tab:rtts_urhi_nriqa}

\resizebox{\columnwidth}{!}{
\begin{tabular}{l c ccc ccc c}
\toprule
\multirow{2}{*}{\textbf{Methods}} &
\multirow{2}{*}{\textbf{Venue}} &
\multicolumn{3}{c}{\textbf{RTTS}} &
\multicolumn{3}{c}{\textbf{URHI}} &
\multirow{2}{*}{\textbf{Latency (ms)$\downarrow$}} \\
\cmidrule(lr){3-5} \cmidrule(lr){6-8}
& &
\textbf{NIQE$\downarrow$} &
\textbf{BRISQUE$\downarrow$} &
\textbf{FADE$\downarrow$} &
\textbf{NIQE$\downarrow$} &
\textbf{BRISQUE$\downarrow$} &
\textbf{FADE$\downarrow$} &
\\
\midrule

CoA~\cite{ma2025coa}
& CVPR'25
& 4.7287 & 30.97 & 0.8585
& 4.0410 & 30.61 & 0.9145
& 179.81 \\

IPC-Dehaze~\cite{fu2025iterativeGAN}
& CVPR'25
& 4.0953 & 19.61 & 1.0984
& 3.8983 & 17.21 & 1.0903
& 1837.11 \\

DiffDehaze~\cite{wang2025diffusion2}
& CVPR'25
& 3.8148 & 16.64 & 1.1324
& 3.8371 & \underline{14.11} & 1.1503
& 6528.29 \\

DehazeSB~\cite{lan2025schrodinger}
& ICCV'25
& 3.8492 & 24.88 & 0.8244
& 3.6002 & 24.33 & 0.7913
& 221.65 \\

HazeFlow~\cite{shin2025hazeflow}
& ICCV'25
& 4.9487 & \textbf{4.73} & \underline{0.5816}
& -- & -- & --
& 485.30 \\

BiLaLoRA~\cite{zhang2026bilevel}
& CVPR'26
& 5.0965 & 24.32 & 0.7515
& 4.1799 & 32.13 & 0.8758
& 185.07 \\

\midrule

\rowcolor{gray!15}
\textbf{PGL-Net-T}
& --
& \underline{3.7458} & 15.29 & 0.5850
& \underline{3.3813} & 14.25 & \underline{0.5392}
& \textbf{14.17} \\

\rowcolor{gray!15}
\textbf{PGL-Net-S}
& --
& \textbf{3.6032} & \underline{8.16} & \textbf{0.4076}
& \textbf{3.1051} & \textbf{11.84} & \textbf{0.4620}
& \underline{26.31} \\

\bottomrule
\end{tabular}
}
\end{table}

\subsection{Comparison with State-of-the-Arts}

\subsubsection{Quantitative Evaluation.}
On the three paired real-world benchmarks in Tables~\ref{tab:rrshid_comparison}--\ref{tab:nhhaze21_comparison}, PGL-Net achieves strong restoration results with substantially lower complexity. In particular, PGL-Net-S surpasses the previous best RRSHID result by 0.99 dB, while the 0.78M-parameter PGL-Net-T remains competitive on RW$^2$AH and spatially non-homogeneous NH-HAZE21 using only 2.71G MACs.

On the unpaired RTTS and URHI benchmarks
(Table~\ref{tab:rtts_urhi_nriqa}), PGL-Net-S achieves the best NIQE
and FADE on both datasets, as well as the best BRISQUE on URHI, while
ranking second on RTTS BRISQUE. Meanwhile, PGL-Net-T maintains competitive
perceptual quality with only 14.17 ms latency, over $12\times$ faster
than the fastest recent baseline. These results demonstrate that PGL-Net achieves a favorable quality--efficiency trade-off for real-world deployment.

\subsubsection{Qualitative Evaluation.}
Visual comparisons on RRSHID and RW$^2$AH are shown in Figs.~\ref{fig:qualitative_rrshid} and~\ref{fig:qualitative_rw2ah}, respectively.
Traditional priors like DCP~\cite{he2009DCP} often suffer from severe color distortion.
While recent deep models like DehazeFormer~\cite{song2023dehazeformer} and SGDN~\cite{fang2025sgdn} improve visibility, they tend to leave residual veils in thick haze regions or over-smooth fine textures.
In contrast, benefiting from the physics-inspired global-local decoupling, our lightweight PGL-Net-T successfully penetrates dense haze and restores faithful colors and sharp structures closest to the ground truth.
More qualitative comparisons and failure cases are provided in the supplementary material.

\begin{figure*}[!t]
  \centering
  \includegraphics[width=1.0\textwidth]{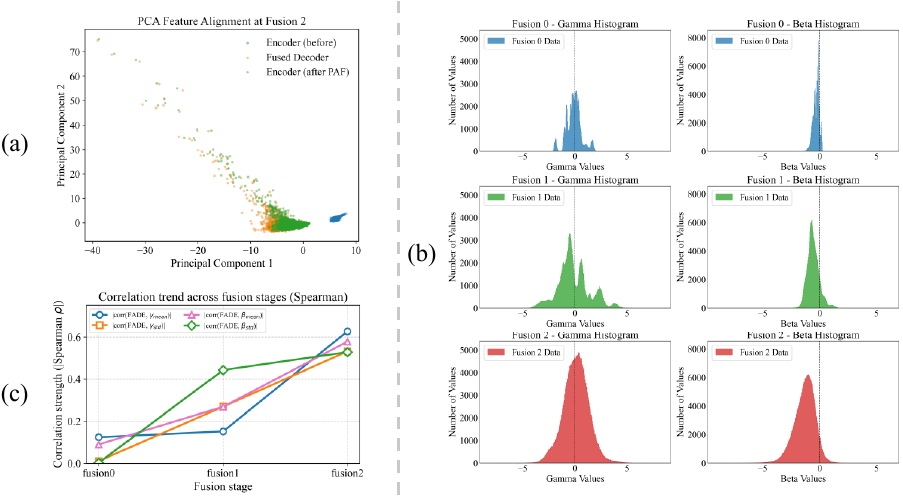}
  \caption{Evidence of degradation-aware affine rectification by PAF on RTTS.
Fusion-0/1/2 denote skip stages from shallow to deep.
(a) PCA visualization shows clear encoder--decoder alignment after PAF (mean distance decreases by 60.1\%, Fr\'echet distance~\cite{heusel2017fid} decreases by 80.6\%).
(b) The bias $\beta$ is dominantly negative and becomes increasingly negative with depth, consistent with $\beta \rightarrow A(1-\frac{1}{t})$, while the variance of $\gamma$ increases.
(c) Spearman correlation with FADE~\cite{choi2015fade} strengthens across stages (from $\approx$0.1 to $\approx$0.6), indicating haze-dependent affine rectification.}
  \label{fig:evidence}
\end{figure*}

\begin{table}[t]
\centering
\caption{Object Detection Performance on the RTTS Dataset}
\label{tab:yolo_detection}
\setlength{\tabcolsep}{2pt}
\resizebox{\columnwidth}{!}{
\begin{tabular}{@{} l c c c c c c c@{}}
\toprule
\textbf{Methods} & \textbf{Person} & \textbf{Bicycle} & \textbf{Car} &
\textbf{Motor} & \textbf{Bus} & \textbf{mAP$_{50:95}$} & \textbf{Latency (ms)}\\
\midrule
Hazy Input & 49.1 & 26.4 & 37.1 & 22.5 & 23.9 & 31.8 & -- \\
DCP \cite{he2009DCP} & 49.2 & 27.0 & 37.4 & 23.7 & 24.1 & 32.3 & -- \\
FSDGN \cite{yu2022fsdgn} & 48.8 & 26.5 & 38.4 & 23.3 & 24.1 & 32.2 & 32.72 \\
DehazeFormer \cite{song2023dehazeformer} & 48.8 & 26.6 & 37.6 & 23.8 & 24.0 & 32.2 & 313.14 \\
DEA-Net \cite{chen2024dea} & 48.2 & 26.0 & 37.0 & 22.9 & 23.3 & 31.5 & 28.23 \\
OKNet \cite{cui2024oknet} & 49.1 & 26.9 & 38.3 & 24.2 & 24.9 & 32.7 & 29.10 \\
WDMamba \cite{sun2025wdmamba} & 48.9 & 27.3 & 38.1 & 24.8 & 24.9 & 32.8 & 82.13 \\
SGDN\cite{fang2025sgdn} & 49.5 & \second{27.5} & 38.4 & 24.5 & \second{25.6} & 33.1 & 146.45 \\
\midrule
\rowcolor{gray!12}
\textbf{PGL-Net-T} & \second{49.6} & \best{27.7} & \second{38.7} & \second{25.2} & 25.3 & \second{33.3} & \best{14.17} \\
\rowcolor{gray!12}
\textbf{PGL-Net-S} & \best{49.8} & 27.2 & \best{39.4} & \best{25.7} & \best{26.2} & \best{33.7} & \second{26.31} \\
\bottomrule
\end{tabular}
}
\end{table}

\subsection{Boosting High-Level Perception}
To verify the utility of dehazing for high-level perception, we assess YOLO26~\cite{sapkota2025yolo26} object detection on RTTS~\cite{li2018rttsAndReside} after applying different dehazing methods, as summarized in Table~\ref{tab:yolo_detection}.
The reported latency is the dehazing inference time.
Raw hazy images achieve 31.8\% mAP$_{50:95}$, while several prior dehazing approaches provide only limited or inconsistent improvements.
In contrast, PGL-Net-S achieves the best performance of 33.7\%, improving over the hazy baseline by 1.9\% and surpassing SGDN~\cite{fang2025sgdn} by 0.6\%, while reducing dehazing latency from 146.45\,ms to 26.31\,ms.
The lightweight PGL-Net-T also delivers competitive accuracy of 33.3\% with a latency of only 14.17\,ms, outperforming heavier models such as DehazeFormer~\cite{song2023dehazeformer} in both
accuracy and efficiency.
Additional SAM-based qualitative comparisons in the supplementary material show that PGL-Net better preserves object boundaries and region coherence on RW$^2$AH.

\subsection{Mechanism Analysis of PAF}
\begin{figure}[t]
  \centering
  \includegraphics[width=\columnwidth]{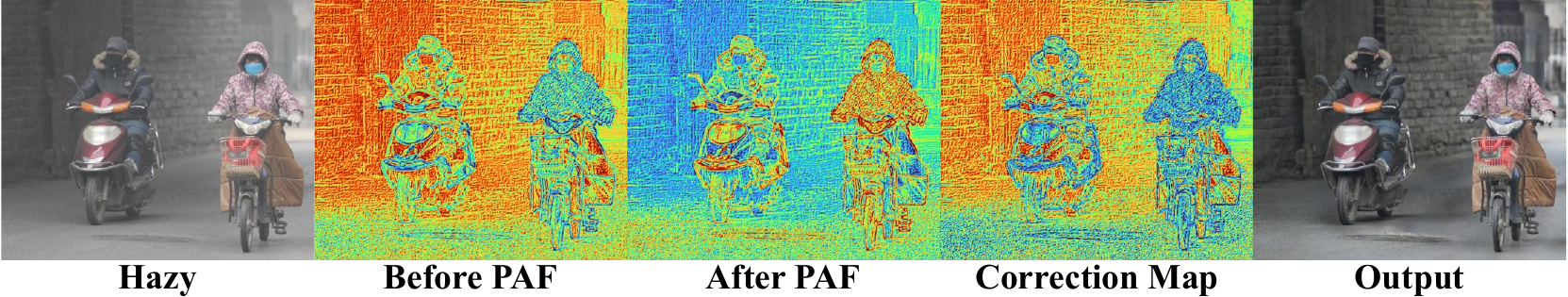}
  \caption{Spatial visualization of PAF responses on RTTS.}
  \label{fig:paf_spatial}
\end{figure}

As shown in Fig.~\ref{fig:evidence}, PAF reduces the mean feature distance by
60.1\% and the Fr\'echet distance~\cite{heusel2017fid} by 80.6\%. The increasingly
negative bias and stronger FADE correlation across fusion stages support
degradation-aware rectification. Fig.~\ref{fig:paf_spatial} further shows
stronger responses in hazy backgrounds than clearer foregrounds, illustrating
content-dependent correction as global affine factors act on spatially varying
features. Together, these results suggest that PAF applies progressively
stronger cross-branch rectification at deeper fusion stages while retaining
spatially varying responses to local haze conditions.

\subsection{Ablation Study}

\begin{table}[!htbp]
\centering
\caption{Ablation study of the proposed modules.}
\label{tab:ablation_module}
\setlength{\tabcolsep}{2.2pt}
\renewcommand{\arraystretch}{1.12}
\resizebox{\columnwidth}{!}{
\begin{tabular}{l ccc ccc cc}
\toprule
\multirow{2}{*}{\textbf{Settings}} &
\multicolumn{3}{c}{\textbf{RW$^2$AH}} &
\multicolumn{3}{c}{\textbf{RRSHID}} &
\multicolumn{2}{c}{\textbf{Overhead}} \\
\cmidrule(lr){2-4} \cmidrule(lr){5-7} \cmidrule(lr){8-9}
& \textbf{PSNR$\uparrow$} & \textbf{SSIM$\uparrow$} & \textbf{LPIPS$\downarrow$}
& \textbf{PSNR$\uparrow$} & \textbf{SSIM$\uparrow$} & \textbf{LPIPS$\downarrow$}
& \textbf{Params} & \textbf{MACs} \\
\midrule
Baseline & 21.46 & 0.5976 & 0.3357 & 23.47 & 0.6696 & 0.4229 & 0.84M & 2.81G \\
+ DAM & 22.25 & 0.6264 & 0.3195 & 24.65 & 0.6977 & 0.4071 & \best{0.74M} & \best{2.71G} \\
+ PAF & 22.45 & 0.6324 & 0.3169 & 24.55 & 0.7039 & 0.3998 & 0.89M & 2.81G \\
\rowcolor{gray!15}
\textbf{+ DAM + PAF} & \best{22.60} & \best{0.6468} & \best{0.3045}
& \best{24.85} & \best{0.7074} & \best{0.3982} & 0.78M & \best{2.71G} \\
\bottomrule
\end{tabular}}
\end{table}

We ablate DAM and PAF in PGL-Net-T on RW$^2$AH and RRSHID
(Table~\ref{tab:ablation_module}). Individually, DAM and PAF improve RW$^2$AH
PSNR from 21.46 to 22.25 and 22.45~dB, respectively. Combining both modules
reaches 22.60~dB with minimal overhead. On RRSHID, the combined setting
improves PSNR from 23.47 to 24.85~dB and reduces LPIPS from 0.4229 to 0.3982,
indicating complementary benefits on the evaluated datasets.

Additional ablations examine the skip-fusion design. With comparable complexity,
PAF outperforms Attention Gate~\cite{oktay2018AG} and
SKFusion~\cite{song2022gUNet,song2023dehazeformer} by 0.66 and 0.35~dB.
It also exceeds the encoder-only (Affine-E) and decoder-only (Affine-D)
variants by 1.07 and 1.74~dB, supporting joint conditioning of both branches
over FiLM-like~\cite{perez2018film} single-branch modulation. Compared with
spatial $\ell_1$ supervision, real--imaginary supervision improves PSNR by
0.29~dB and reduces LPIPS by 0.0178. See the supplementary material for full results.
\section{Conclusion}

We propose PGL-Net, a lightweight physics-inspired framework that decouples
dehazing into global distribution rectification and local structural refinement.
By integrating PAF into skip connections and DAM into the backbone, PGL-Net
embeds physical priors while remaining highly efficient. Extensive experiments
demonstrate state-of-the-art real-world dehazing performance and improved
downstream detection at low latency, making it well-suited for practical edge deployment.

\clearpage

{
    \small
    \bibliographystyle{ieeenat_fullname}
    \bibliography{main}
}

\end{document}


\raggedbottom

\twocolumn[{
\begin{center}
    {\LARGE\bfseries Efficient Real-World Dehazing via\par}
    \vspace{0.15em}
    {\LARGE\bfseries Physics-Inspired Global-Local Decoupling\par}
    \vspace{0.8em}
    {\large\bfseries Supplementary Material\par}
    \vspace{0.8em}
    {\large Yifei Qu \quad Ru Li$^\dagger$ \quad Junjie Chen \quad Jinyuan Wu\par}
    \vspace{0.25em}
    {\normalsize Harbin Institute of Technology, China\par}
    {\tt\small 2023211862@stu.hit.edu.cn, liru@hit.edu.cn, csjunjie.c@gmail.com, 25s130347@stu.hit.edu.cn\par}
\end{center}

\vspace{0.8em}
}]
\begingroup
\renewcommand{\thefootnote}{\fnsymbol{footnote}}
\footnotetext[2]{Corresponding author.}
\endgroup

\begin{figure*}[t!]
    \centering
    \includegraphics[width=\textwidth]{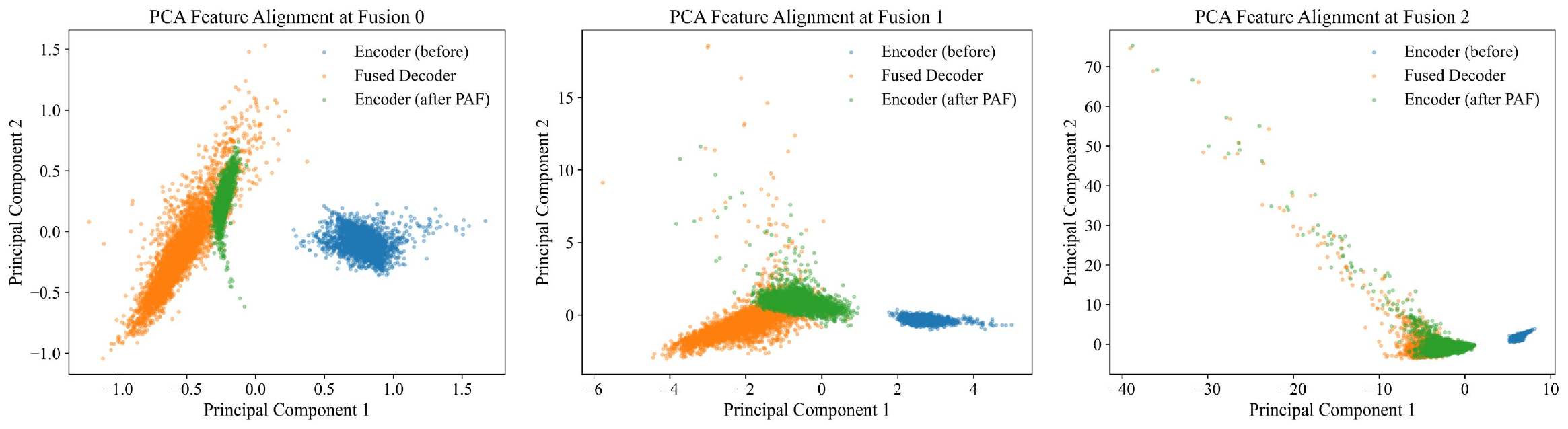}
    \caption{PCA visualization of encoder and decoder feature distributions
    across the three fusion stages. Encoder features before fusion, decoder
    features, and encoder features after PAF rectification are shown in blue,
    orange, and green, respectively.}
    \label{fig:pca_all}
\end{figure*}

This supplementary material provides additional methodological and
experimental details for PGL-Net. We first describe the dataset protocols and
training settings in Sec.~\ref{sec:data_impl}. Sec.~\ref{sec:paf_analysis} then
extends the PAF mechanism analysis in the main paper with enlarged multi-stage
feature visualizations, detailed interpretation of the learned affine factors,
and further discussion of the spatial correction responses shown in the main
paper (Fig.~\ref{fig:pca_all}). We further provide a visual
comparison on NH-HAZE21 (Fig.~\ref{fig:nhhaze21_qualitative}) and investigate
how different dehazing results affect SAM-based segmentation in
Sec.~\ref{sec:segmentation}. Additional results include analyses of
skip-connection strategies and frequency-domain supervision
(Sec.~\ref{sec:additional_ablation}), latency on multiple hardware platforms
(Sec.~\ref{sec:latency}), representative failure cases and limitations
(Sec.~\ref{sec:limitations}), and qualitative comparisons on RRSHID and RTTS
(Sec.~\ref{sec:qualitative}).

\appendix

\section{Dataset and Implementation Details}
\label{sec:data_impl}

\subsection{Datasets}

\textbf{RRSHID~\cite{zhu2025RRSHID}.}
The Real-World Remote Sensing Hazy Image Dataset is a large-scale paired
benchmark containing 3,053 hazy and haze-free remote-sensing image pairs.
It covers urban, agricultural, and coastal scenes under diverse atmospheric
conditions. The dataset is divided into thin-, moderate-, and thick-haze
subsets according to visibility, allowing the restoration performance to be
examined across different degradation severities. Following the official
split, 2,749 pairs are used for training and 304 pairs for testing.

\textbf{RW$^2$AH~\cite{fang2025sgdn}.}
The Real-world Well-Aligned Haze dataset contains 1,758 paired images captured
by stationary webcams across diverse geographic regions and scene types.
Hazy and clear observations are acquired from the same viewpoint within a
short time interval, which improves spatial alignment while retaining
realistic illumination and weather variation. We use the official split of
1,406 training pairs and 352 testing pairs.

\textbf{NH-HAZE21~\cite{ancuti2021nh}.}
NH-HAZE21 is a paired real-world benchmark introduced in the NTIRE 2021
NonHomogeneous Dehazing Challenge. It contains 35 pairs of real outdoor hazy
and haze-free images, with spatially varying haze generated using professional
haze machines. We use 30 pairs for training and the remaining 5 pairs for
testing. Its non-uniform haze distribution supports evaluation under locally
varying degradation rather than only globally homogeneous scattering.

\textbf{RTTS~\cite{li2018rttsAndReside}.}
The Real-world Task-driven Testing Set contains 4,322 outdoor hazy images with
object bounding-box annotations. It includes traffic-related categories such
as people, bicycles, cars, buses, and motorbikes. In addition to the detection
evaluation reported in the main paper, we use RTTS to examine visual behavior
under dense haze, distant scenes, and reduced local contrast.

\textbf{URHI~\cite{li2018rttsAndReside}.}
The Unannotated Real Hazy Images subset of RESIDE contains 4,807 real-world
hazy images collected from unconstrained outdoor scenes. It covers diverse
haze densities, illumination conditions, and scene contents, but does not
provide paired haze-free ground truth or task annotations. We therefore use
URHI for no-reference perceptual evaluation with NIQE, BRISQUE, and FADE, as
reported in the main paper. Together, RTTS and URHI assess restoration behavior
on real hazy observations beyond paired full-reference benchmarks.

\subsection{Implementation Details}

PGL-Net follows a symmetric U-shaped architecture. The numbers of
Degradation-Aware Modulation blocks across the seven stages are
$\{M,M,M,2M,M,M,M\}$, and the corresponding channel dimensions are
$\{N,2N,4N,8N,4N,2N,N\}$. Both PGL-Net-T and PGL-Net-S use a depthwise
convolution kernel size of $5$ and a base channel width of $N=24$. Their block
repetition factors are $M=2$ and $M=4$, respectively.

We optimize the models with AdamW using $\beta_1=0.9$ and $\beta_2=0.999$.
The learning rate follows a cosine annealing schedule from $4\times10^{-4}$
to $4\times10^{-8}$. Batch sizes are 64 for PGL-Net-T and 32 for PGL-Net-S.
Training images are randomly cropped into $256\times256$ patches and augmented
with horizontal flips and $90^\circ$ rotations. Both variants are trained for
800 epochs on RRSHID. On RW$^2$AH and NH-HAZE21, PGL-Net-T and PGL-Net-S are
trained for 300 and 600 epochs, respectively. All models are implemented in PyTorch and
trained on a single NVIDIA RTX 3090 GPU.

\section{Additional Analysis of Physics-Inspired Affine Fusion}
\label{sec:paf_analysis}

PAF performs globally conditioned, channel-wise affine rectification between
encoder and decoder features. To examine its behavior beyond a single fusion
stage, we project globally pooled features from all three skip connections
into two-dimensional PCA spaces. The stages are ordered from shallow to deep
and denoted Fusion-0, Fusion-1, and Fusion-2. The mechanism-analysis figure in
the main paper (Fig.~4) summarizes the PCA distributions, affine-parameter
statistics, and their correlations with haze severity. The following analysis
provides an enlarged PCA view, detailed interpretation, and further discussion
of the spatial visualization in the main paper.

\textbf{(a) Feature Distribution Alignment.}
Fig.~\ref{fig:pca_all} enlarges the PCA results summarized in Fig.~4(a) of the
main paper. Encoder and decoder features occupy
noticeably different regions before rectification, reflecting a distribution
gap between representations produced at different stages of restoration.
After PAF, the transformed encoder features shift
toward the decoder distribution at all three stages. At Fusion-2, PAF reduces
the mean feature distance by 60.1\% and the Fréchet distance~\cite{heusel2017fid}
by 80.6\%. The alignment becomes progressively clearer in deeper stages, whose
features contain more degradation-sensitive structural and semantic statistics.

\textbf{(b) Affine Parameter Distributions.}
As shown in Fig.~4(b) of the main paper, the learned affine factors exhibit
stage-dependent patterns. The
bias $\beta$ is predominantly negative ($>90\%$ across stages) and becomes
progressively more negative with depth
($-0.30 \rightarrow -0.57 \rightarrow -1.31$). This trend qualitatively agrees
with the inverse-ASM motivation, where haze removal involves an additive bias
correction. Meanwhile, the variance of $\gamma$ increases in deeper stages,
suggesting stronger feature-wise compensation for high-level haze-induced
attenuation. Here, $\gamma$ and $\beta$ are feature-space rectification factors,
not explicit estimates of physical transmission or atmospheric light.

\textbf{(c) Correlation with Haze Severity.}
Fig.~4(c) of the main paper reports the Spearman correlations between haze severity,
measured by FADE~\cite{choi2015fade}, and the learned affine statistics. The
correlation grows from weak ($\approx0.1$) at Fusion-0 to strong ($\approx0.6$)
at Fusion-2. In particular, $\beta$ exhibits a monotonic negative relationship
with haze severity: heavier haze corresponds to more negative bias correction.
This suggests that PAF adapts its rectification to degradation severity.

\textbf{(d) Spatial Correction Visualization.}
Although PAF generates global channel-wise affine factors, their application
to spatially varying encoder features yields the non-uniform correction
$\Delta F_c(x)=F_{\mathrm{enc},c}(x)(\gamma_c-1)+\beta_c$.
As shown in Fig.~7 of the main paper, PAF responds more strongly to hazy,
low-contrast backgrounds and more weakly to relatively clear foregrounds.
Thus, PAF performs content-dependent rectification rather than uniform global
enhancement and complements DAM's fine-grained handling of locally varying
degradation.

\begin{figure*}[t]
    \centering
    \includegraphics[width=\textwidth]{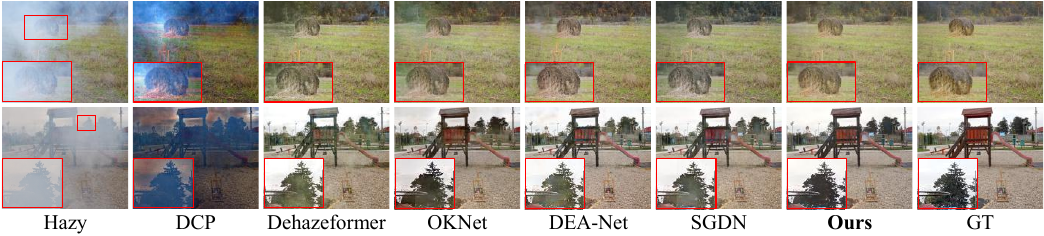}
    \caption{Visual comparison on NH-HAZE21. The spatially non-homogeneous
    haze provides a complementary qualitative evaluation to the quantitative
    results reported in the main paper. Zoom in for a better view.}
    \label{fig:nhhaze21_qualitative}
\end{figure*}

\section{Impact of Dehazing on Image Segmentation}
\label{sec:segmentation}

Beyond the object-detection results in the main paper, we examine whether
dehazing also benefits segmentation-oriented perception. We apply the Segment
Anything Model (SAM)~\cite{kirillov2023sam} to images restored by different
dehazing methods and compare the resulting segmentation maps on
RW$^2$AH~\cite{fang2025sgdn}.

\begin{figure*}[t]
    \centering
    \includegraphics[width=\textwidth]{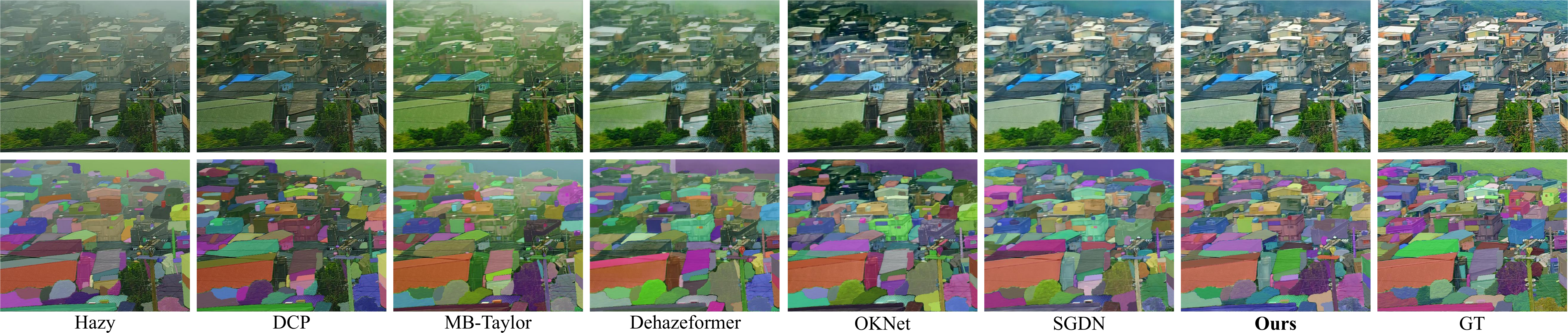}
    \caption{Qualitative comparison of dehazing results and corresponding SAM
    segmentation maps on RW$^2$AH. The top row shows restored images produced
    by different dehazing methods, while the bottom row shows the corresponding
    SAM segmentation results.}
    \label{fig:segmentation}
\end{figure*}

As shown in Fig.~\ref{fig:segmentation}, hazy inputs often produce fragmented
regions and inaccurate boundaries because haze suppresses contrast and
obscures structures. Some restoration methods improve segmentation consistency
but may introduce color distortion or over-smoothed details. In contrast,
images restored by PGL-Net lead to clearer boundaries and more coherent
regions, suggesting that the proposed restoration process can also benefit
segmentation-oriented perception.

\section{Additional Ablation Studies}
\label{sec:additional_ablation}

The main paper reports the core component ablation across RW$^2$AH and
RRSHID. Here, we further examine alternative skip-connection strategies,
frequency-domain supervision, and the qualitative effects of DAM and PAF.

\begin{figure}[H]
    \centering
    \includegraphics[width=\columnwidth]{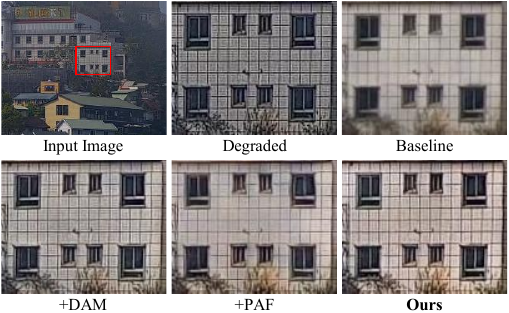}
    \caption{Qualitative ablation of the proposed modules. DAM improves local
    structures, while PAF further corrects global contrast and residual haze.}
    \label{fig:ablation}
\end{figure}

\begin{table}[t]
\centering
\caption{Comparison of different skip-connection strategies on RW$^2$AH.}
\label{tab:ablation_skip}
\setlength{\tabcolsep}{3pt}
\renewcommand{\arraystretch}{1.12}
\resizebox{\columnwidth}{!}{
\begin{tabular}{l cccc}
\toprule
\textbf{Fusion Strategy} &
\textbf{PSNR$\uparrow$} &
\textbf{SSIM$\uparrow$} &
\textbf{LPIPS$\downarrow$} &
\textbf{MACs} \\
\midrule
Summation
& 21.96 & 0.6270 & 0.3171 & \best{2.71G} \\

Concatenation
& 21.90 & 0.6233 & 0.3205 & 2.93G \\

Attention Gate~\cite{oktay2018AG}
& 21.94 & 0.6220 & 0.3212 & 3.06G \\

SKFusion~\cite{song2023dehazeformer,song2022gUNet}
& 22.25 & 0.6264 & 0.3195 & \best{2.71G} \\

\midrule
Affine-E
& 21.53 & 0.6112 & 0.3317 & \best{2.71G} \\

Affine-D
& 20.86 & 0.6067 & 0.3584 & \best{2.71G} \\

\rowcolor{gray!15}
\textbf{PAF (E+D)}
& \best{22.60} & \best{0.6468} & \best{0.3045} & \best{2.71G} \\
\bottomrule
\end{tabular}}
\end{table}

PAF achieves the best restoration quality among all evaluated fusion
strategies while retaining the same 2.71G MACs as summation and SKFusion.
This comparison suggests an advantage of the ASM-motivated scale--shift
rectification over direct or attention-based skip fusion.
To isolate the effect of joint conditioning from the affine operator itself,
we further construct two single-branch variants, Affine-E and Affine-D,
which retain the same scale--shift rectification but derive the affine
factors from only the encoder or decoder descriptor, respectively.
PAF improves PSNR over Affine-E and Affine-D by 1.07 and 1.74~dB,
respectively, with nearly identical model size (0.78M vs.\ 0.77M parameters).
Together, these comparisons indicate that PAF benefits from both the
ASM-motivated operator structure and joint encoder--decoder conditioning.
The gains over Affine-E and Affine-D specifically show that joint conditioning,
rather than a generic FiLM-like~\cite{perez2018film} affine operator alone,
is important for effective skip rectification.

\begin{table}[t]
\centering
\caption{Ablation of frequency-domain loss on RW$^2$AH.}
\label{tab:ablation_fft}
\resizebox{\columnwidth}{!}{
\begin{tabular}{l ccc}
\toprule
\textbf{Loss Configuration} & \textbf{PSNR$\uparrow$} & \textbf{SSIM$\uparrow$} & \textbf{LPIPS$\downarrow$} \\
\midrule
Spatial L1 Only & 22.31 & 0.6146 & 0.3223 \\
\rowcolor{gray!15}
\textbf{+ Real/Imag} & \best{22.60} & \best{0.6468} & \best{0.3045} \\
\bottomrule
\end{tabular}}
\end{table}

Compared with spatial reconstruction alone, real--imaginary supervision
consistently improves PSNR, SSIM, and LPIPS, indicating that direct
complex-domain supervision provides a stable frequency constraint.

\section{Latency on Different Hardware Platforms}
\label{sec:latency}

In addition to the fair PyTorch FP32 latency comparison with existing methods
in the main paper, we conduct a separate FP16 deployment study of PGL-Net on
representative computing devices with $512\times512$ inputs. GPU inference is
accelerated using TensorRT, while CPU inference is optimized with OpenVINO.

\begin{table}[t]
\centering
\caption{Inference latency (ms) at FP16 precision on representative devices.}
\label{tab:latency}
\resizebox{\columnwidth}{!}{
\begin{tabular}{l ccc}
\toprule
\textbf{Method} & \textbf{Tesla T4} & \textbf{RTX 3090} & \textbf{CPU} \\
\midrule
PGL-Net-T & 12.90 & 3.92 & 14.69 \\
PGL-Net-S & 23.11 & 6.68 & 24.58 \\
\bottomrule
\end{tabular}
}
\end{table}

As reported in Table~\ref{tab:latency}, PGL-Net-T achieves real-time
performance across all devices, running at 3.92\,ms on RTX 3090, 12.90\,ms on
Tesla T4, and 14.69\,ms on an Intel Xeon Gold 6459C CPU. PGL-Net-S remains efficient, with latency below
7\,ms on RTX 3090 and 23.11\,ms on Tesla T4. These results demonstrate that
PGL-Net is readily deployable with common inference backends. Training and
evaluation code, pretrained weights, Python/C++ inference implementations, and
cross-backend latency scripts will be released upon publication.

\section{Limitations}
\label{sec:limitations}

\begin{figure}[H]
    \centering
    \includegraphics[width=\columnwidth]{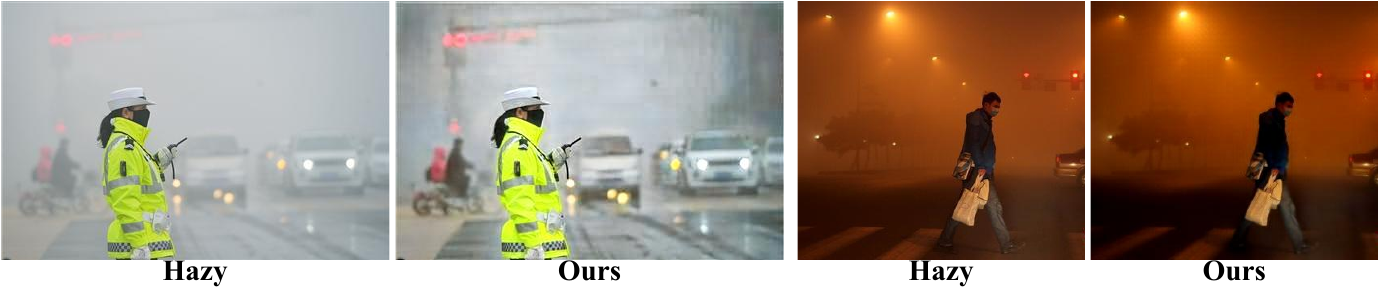}
    \caption{Failure cases. PGL-Net may produce incomplete haze removal or
    color bias under extremely dense haze and nighttime scenes with complex
    local illumination.}
    \label{fig:failure}
\end{figure}

PGL-Net is designed for a favorable balance between restoration quality and
deployment efficiency. Under extremely dense haze, however, scene information
may be irreversibly attenuated before restoration, resulting in incomplete haze
removal or over-smoothed details. Nighttime scenes with colored local light
sources can also deviate from the simplified atmospheric-light assumption and
lead to local color bias or unnatural contrast. In addition, PAF generates
channel-wise affine factors from global statistics. Although their interaction
with spatially varying features produces content-dependent correction, highly
localized degradation may still require stronger spatial priors. Future work
will investigate flexible illumination modeling and efficient spatially aware
conditioning for these challenging cases.

\section{Additional Qualitative Comparisons}
\label{sec:qualitative}

We conclude with additional qualitative comparisons on RRSHID and RTTS.
These examples complement the main figures with additional haze densities,
scene layouts, and object scales.

\begin{figure*}[t]
    \centering
    \includegraphics[width=\textwidth]{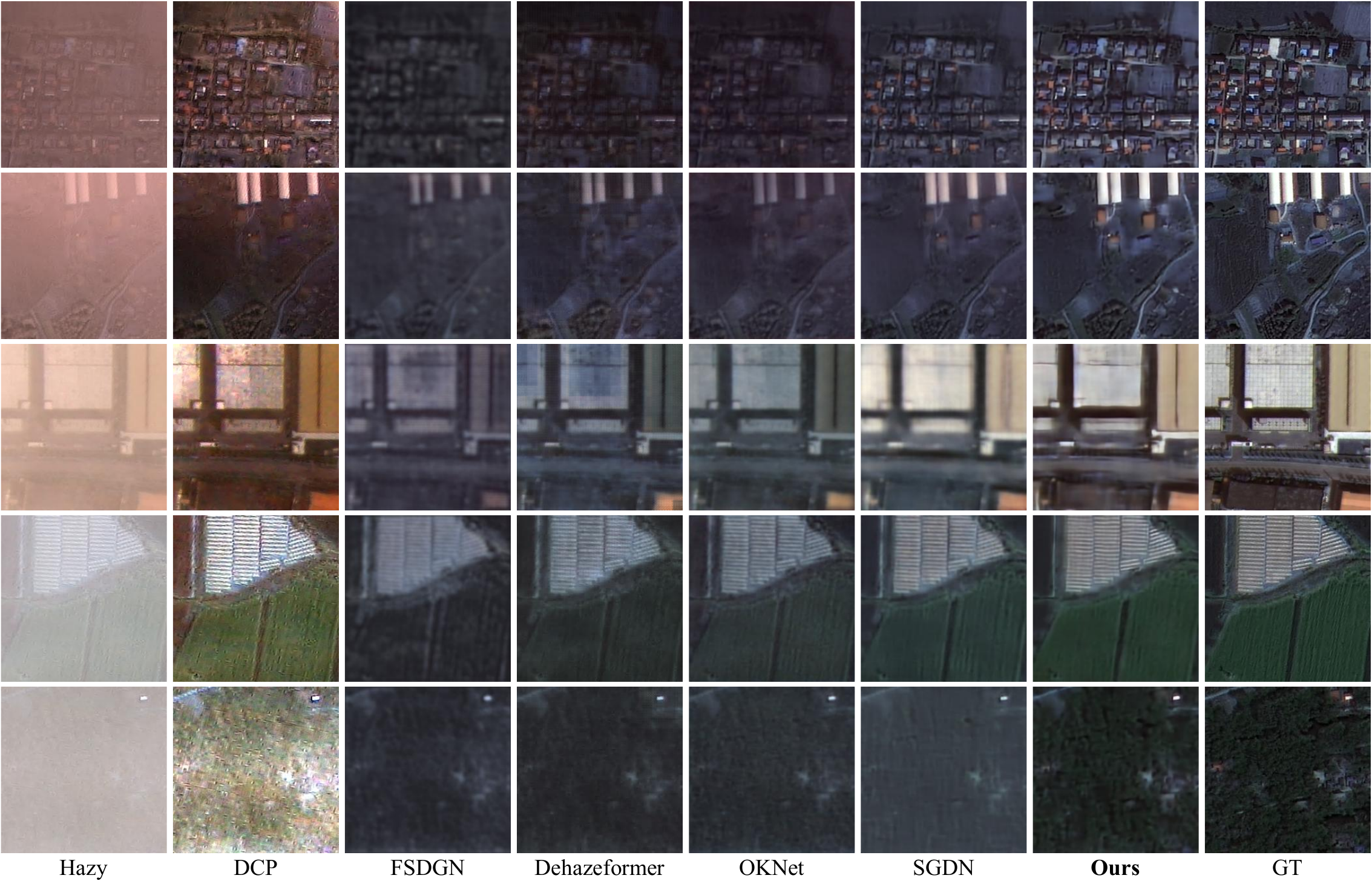}
    \caption{Additional visual comparison on RRSHID. The proposed method
    restores natural contrast and color while retaining fine structures under
    different haze densities. Zoom in for a better view.}
    \label{fig:rrshid_supplement}
\end{figure*}

Fig.~\ref{fig:rrshid_supplement} shows paired examples from RRSHID. Traditional
priors may introduce color distortion, while several learning-based methods
leave residual haze or suppress fine textures. PGL-Net produces cleaner
results with more natural color and better-preserved structural details.

\begin{figure*}[t]
    \centering
    \includegraphics[width=\textwidth]{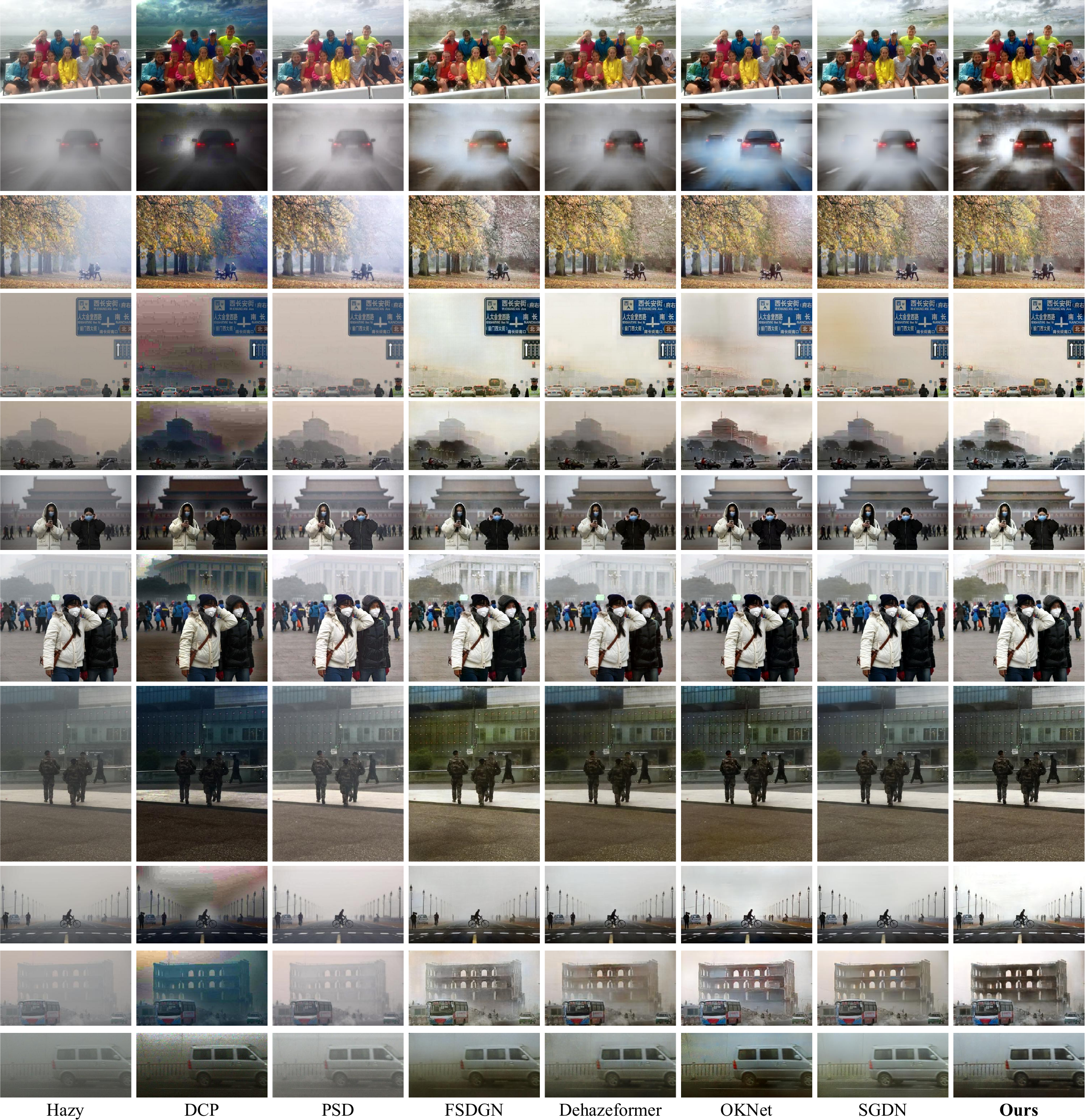}
    \caption{Additional visual comparison on RTTS. PGL-Net improves visibility
    in distant and densely hazy regions while preserving objects and scene
    structure. Zoom in for a better view.}
    \label{fig:rtts_supplement}
\end{figure*}

Fig.~\ref{fig:rtts_supplement} presents unpaired real-world traffic scenes from
RTTS. Such images contain dense fog, distant haze, low contrast, and small
objects. The proposed global--local design removes broad haze veils while
retaining local boundaries, yielding visually consistent inputs for the
downstream detection evaluation reported in the main paper.

\FloatBarrier
{
    \small
    \bibliographystyle{ieeenat_fullname}
    \bibliography{main}
}